# Modus Ponens Generating Function in the Class of $\wedge$-valuations of Plausibility

**Ildar Z. Batyrshin**
Kazan State Technological University
K. Marx st. 68, Kazan, 420015
TATARSTAN, RUSSIA

## Abstract

We discuss the problem of construction of inference procedures which can manipulate with uncertainties measured in ordinal scales and fulfil to the property of strict monotonicity of conclusion. The class of $\wedge$-valuations of plausibility is considered where operations based only on information about linear ordering of plausibility values are used. In this class the modus ponens generating function fulfiling to the property of strict monotonicity of conclusions is introduced.

## 1 STABILITY OF DECISIONS IN INFERENCE PROCEDURES

Human judgements about plausibility, truth, certainty values of premises, rules and facts are usually qualitative and measured in ordinal scales. Representation of these judgements by numbers from interval $L = [0,1]$ or $L = [0,100]$ and using over these numbers quantitative operations such as multiplication, addition and so on is not always correct. Let's consider example.

Let $R_1$ and $R_2$ are two rules of some expert system:

$$R1: \text{ If } A_1 \text{ then } H_1, pv(R_1), \tag{1}$$

$$R2: \text{ If } A_2 \text{ then } H_2, pv(R_2), \tag{2}$$

where $pv(R_1)$ and $pv(R_2)$ are the plausibility, certainty, truth values of rules measured in some linearly ordered scale $L$, for example $L = [0,1]$. Often plausibilities of conclusions are calculated by:

$$pv(H_1) = pv(R_1) * pv(A_1), \tag{3}$$

$$pv(H_2) = pv(R_2) * pv(A_2), \tag{4}$$

where $pv(A_1)$ and $pv(A_2)$ are the plausibilities of premises and $*$ is some $T$-norm, for example multiplication operation (Godo, Lopez de Mantaras et al. 1988; Hall 1990; Trillas, Valverde 1985; Valverde, Trillas 1985; Forsyth 1984). Generally the plausibility of conclusion can be calculated by means of a modus ponens generating function mpgf:

$$pv(H_1) = \text{mpgf}(pv(A_1), pv(R_1)).$$

Let in (1)-(4) the qualitative information about plausibility values is the next:

$$pv(A_1) < pv(A_2) < pv(R_2) < pv(R_1), \tag{5}$$

that is the plausibility values of premises are less than the plausibility values of rules, the plausibility value of $A_1$ is less than the plausibility value of $A_2$ and the plausibility value of rule $R2$ is less than the plausibility value of rule $R1$. Let these plausibility values are interpreted as the next quantitative values from $L = [0,1]$:

$$pv(A_1) = 0.3 < pv(A_2) = 0.4 < pv(R_2) = 0.6 <$$

$$< pv(R_1) = 0.9.$$

If in (3), (4) the operation $*$ will be a multiplication operation then we will obtain from (3)-(4):

$$pv(H_1) = 0.27 > pv(H_2) = 0.24.$$

If the plausibility values from (5) will obtain some another quantitative values preserving the qualitative relation (5), for example $pv(A_1)$ will be changed to $pv(A_1) = 0.2$, then we will obtain the opposite ordering of conclusions:

$$pv(H_1) = 0.18 < pv(H_2) = 0.24.$$

Thus the small transformations in quantitative interpretation of judgements of experts and expert systems users which preserve the qualitative information about plausibility values can bring to opposite results on the output of expert system. The similar situation of instability of results on the output of inference procedure can also arise when we use another quantitative operations in (3)-(4) and such instability of decisions can arise on the each step of inference procedure.

Stability of decisions on the output of inference procedures is achieved for uncertainties measured in ordinal scales if in (3)-(4) we use $* = \wedge(\min)$ operation (Zadeh 1965). But in this case the property of strict monotonicity of conclusions desirable for inference procedures does not fulfiled. This property is discussed in the next section.

## 2 STRICT MONOTONICITY OF CONCLUSIONS

In inference procedures we can also require the fulfilment of the next property of strict monotonicity of conclusions for rules (1) - (2):

*SMC.* If $pv(R1) = pv(R2) > 0$ and $pv(A_1) > pv(A_2)$

then $pv(H_1) > pv(H_2)$.

This requirement is fulfiled for the most quantitative $T$-norms used in (3), (4) instead of $*$ operation but it is not fulfiled for $* = \wedge$ . For example if $pv(R1) = pv(R_2) = 0.6$ and $pv(A_1) = 1 > pv(A_2) = 0.8$ then $pv(H_1) = 0.6 \wedge 1 = 0.6$, $pv(H_2) = 0.6 \wedge 0.8 = 0.6$, i.e. $pv(H_1) = pv(H_2)$ and SMC does not fulfiled.

Here we consider one class of lexicographic valuations of plausibility (Batyrshin 1993, 1994; Batyrshin, Zakuanov 1993) whith lexicographic generalization of $\wedge$ operation. The operations over lexicographic valuations of plausibility are based only on ordinal information about its operands and oriented to manipulation with uncertainties measured in ordinal scale. In this class the modus ponens generating function fulfiling to the property of strict monotonicity of conclusions is introduced.

The concept of lexicography is widely used in combinatorics, informatics and decision-making theory. Algorithms of lexicographic sortings of strings can be find in (Aho, Hopcroft, Ullman 1976; Reingold, Nievergelt, Deo 1977), lexicographically ordered utility functions are considered in (Fishburn 1970), lexicographic ordering of criteria is considered in (Keeney, Raiffa 1976; Podinovskii, Gavrilov 1975), leximin and leximax orderings of criterial values are considered in (Podinovskii, Gavrilov 1975; Moulin 1988; Vilkas 1990). In fuzzy theory, fuzzy and non-monotonic logic leximin and leximax orderings of the weights of the justifications, weights of the formulas, satisfaction degrees and maximal consistent subsets are considered in (Dubois, Lang, Prade 1990; Dubois, Lang, Prade 1992; Benferhat, Cayrol, Dubois, Lang, Prade 1993; Fargier, Lang, Schiex 1993), leximin and leximax ordering of plausibility, certainty values in expert systems and degrees of membership of fuzzy sets are considered in (Batyrshin 1989, 1990, 1993, 1994). The concept of weighted leximin ordering of fuzzy criteria is considered in (Batyrshin, Zakuanov 1990, 1993)

In (Batyrshin 1989, 1990, 1993, 1994; Batyrshin, Zakuanov 1993 ao) a new approach to use the idea of lexicography was developed: in addition to usual lexicographic orderig of strings and lexicographic ordering of ordered strings of values the generalization of main fuzzy logic connectives over such strings was introduced. Different types of such generalization can be developed. Below the most simplest algebra of lexicographic valuations of plausibility is considered.

## 3 $\wedge$-VALUATIONS OF PLAUSIBILITY

Let $L$ is a scale of plausibility (certainty, possibility, truth) values linearly ordered by relation $\leq$ (or $<$ ). For example $L = [0,1], L = [0,100], L = \{a_1 < a_2 < \ldots < a_m\}$, where $a_k$ - some verbal or numerical grades: $L$ = {impossible < almost impossible < slightly possible < average possibility < very possible < almost sure < sure}, $L = \{0,1,2,3,4,5,6\}$, and so on. On such scales min ($\wedge$) and max ($\vee$) operations are defined by relation $\leq$ in the usual way:

$$f \wedge g = f, \; f \vee g = g \text{ if } f \leq g. \tag{6}$$

0 and $I$ will denote the minimal and the maximal elements of $L$. The negation operation $'$ also can be easily introduced by: $f' = I - f$ for quantitative scales $L = [0,1]$ or $L = [0,100]$, and by: $a'_m = a_{m-k+1}$ for symmetric discrete scales.

Let $F$ be a set of all strings of finite length defined on $L$. **An equality relation** $=$ for strings $f(n) = f_1 \ldots f_n$ and $g(m) = g_1 \ldots g_m$ from $F$ where $f_j, g_k \in L$, $j \in J(n) = \{1,\ldots,n\}$, $k \in J(m) = \{1,\ldots,m\}$ is defined as:

$$f(n) = g(m), \text{ if } n = m \text{ and } f_j = g_j \text{ for all } j \in J(n).$$

**A concatenation** of strings $f(n), g(m) \in F$ is a string $fg = f_1 \ldots f_n g_1 \ldots g_m$. **A $\wedge$-string** $f\hat{}$ is a string $f\hat{}(n) = (f_1 \ldots f_n)$ such that $f_j \leq f_{j+1}$ for all $j \in J(n-1)$ if $n > 1$ and $f\hat{}(n) = (f_1)$ if $n = 1$.

**A $\wedge^*$ operation** on $F$ is a sorting operation such that $\wedge^* f$ is a $\wedge$-string obtained from the string $f = f_1 \ldots f_n$ by some permutation of its members.

Let $F\hat{}$ be the set of all $\wedge$-strings from $F$ and $f\hat{}(n)$, $g\hat{}(m)$ $(n,m \geq 1)$ are arbitrary $\wedge$-strings from $F\hat{}$. **An indistinguishability relation $\cong$ on plausibility of $\wedge$-strings** is defined on $F\hat{}$ in the next way:

$f\hat{}(n) \cong g\hat{}(m)$ if $f_1 = g_1 = 0$ or if $f_j = g_j$ for all $j \in J(n \wedge m)$ and $f_{m+1} = I$ if $n > m$ or $g_{n+1} = I$ if $m >$n.

$\wedge$-string is considered as the (intermediate) result of generalized conjunction operation $\Delta$ used over operands $f_1, \ldots, f_n$ and indistinguishability relation $\cong$ generalizes the next identities from the fuzzy logic:

$$a \wedge 0 = 0, \; a \wedge 1 = a.$$

**A reduction operation** $\tilde{}$ on $F\hat{}$ is an operation such that $f\hat{}\tilde{}$ will be the $\wedge$-string with the least length which is indistinguishable with $f\hat{}$. If $f\hat{} = f\hat{}\tilde{}$ then $f\hat{}$ will be called **a $\wedge$-valuation**.

We can show (Batyrshin 1994) that indistinguishability relation $\cong$ on plausibility of $\wedge$-strings is an equivalence relation on $F\hat{}$ and $\wedge$-valuations can be considered as representatives of equivalence classes of this relation. We have $f\hat{}\tilde{} \cong f\hat{}$. For $\wedge$-valuations indistinguishability relation $\cong$ coincides with equality relation $=$. For these reasons we will replace $\wedge$-strings $f\hat{}$ by $\wedge$-valuations $f\hat{}\,\tilde{}$ and below will consider only $\wedge$-valuations. Below $f\hat{}$ will denote $\wedge$-valuation and $F\hat{}$ will denote the set of all $\wedge$-valuations on L.

**An ordering relation $\leq$ on plausibility of $\wedge$-valuations** is defined on $F\hat{}$ in the next way:

$f\hat{}(n) \leq g\hat{}(m)$ if $n \geq m$ and $f_j = g_j$ for all $j \in J(m)$ or if there exist $j \in J(n \wedge m)$ such that $f_j < g_j$ and $f_k = g_k$ for all $k$ <j.

This ordering relation can be considered as generalization of the usual lexicographic ordering relation and the next relation from the fuzzy logic:

$$a \wedge b \leq a.$$

The defined relation $\leq$ will be a linear ordering relation on $F\hat{}$(Batyrshin 1994) and this relation defines on $F\hat{}$ by (6) min ($\wedge$) and max ($\vee$) operations. The elements:

$$0\hat{} = (0), \; I\hat{} = (I),$$

will be accordingly **the minimal** and **the maximal elements** of $F\hat{}$.

Generalizations of **conjunction** $\triangle$ and **disjunction** $\bigtriangledown$ **operations** on $F\hat{}$ are defined in the next way (Batyrshin, Zakuanov 1993, Batyrshin 1993):

$$f\hat{} \triangle g\hat{} = (\wedge^*(f\hat{}g\hat{}))\tilde{},$$

$$f\hat{} \bigtriangledown g\hat{} = f\hat{} \vee g\hat{}.$$

THEOREM 1 (Batyrshin, Zakuanov 1993, Batyrshin 1994). $\triangle$ is a $T$-norm and $\bigtriangledown$ is a $T$-conorm on $F\hat{}$ that is they are commutative, associative, non-decreasing in each argument and fulfil to the properties:

$$f\hat{} \triangle I\hat{} = f\hat{}, \qquad f\hat{} \triangle 0\hat{} = 0\hat{}, \tag{7}$$

$$f\hat{} \bigtriangledown 0\hat{} = f\hat{}, \qquad f\hat{} \bigtriangledown I\hat{} = I\hat{}, \tag{8}$$

$$f\hat{} \triangle g\hat{} < f\hat{} \text{ if } f\hat{} > 0\hat{},\ g\hat{} < I\hat{}, \tag{9}$$

and $\triangle$ is strict monotonic on $F\hat{}$:

$$\text{if } f\hat{} < g\hat{} \text{ and } 0\hat{} < h\hat{} \text{ then } f\hat{} \triangle h\hat{} < g\hat{} \triangle h\hat{}. \tag{10}$$

## 4 NEGATION OPERATION ON $F\hat{}$

Let $' : L \to L$ is a negation operation on $L$ that is for all $f$ from $L$ holds (Weber 1983; Trillas, Alsina, Valverde 1982): $f' \leq g'$ if $g \leq f$; $0' = I$, $I' = 0$. $'$ is a weak negation on $L$ if $f'' \geq f$, and $'$ is an involution on $L$ if $f'' = f$.

A function $c : F\hat{} \to F\hat{}$ will be called **a negation operation** on $F\hat{}$ iff on $F\hat{}$ holds:

$$c(0\hat{}) = I\hat{},\ c(I\hat{}) = 0\hat{},$$

$$c(f\hat{}) \leq c(g\hat{}) \text{ if } g\hat{} \leq f\hat{},$$

$$c(f\hat{}(n)) = (f_1') \text{ if } n = 1.$$

THEOREM 2 (Batyrshin, Zakuanov 1993; Batyrshin 1994). The function $c : F\hat{} \to F\hat{}$ defined by: $c(f\hat{}) = (f_1')$ is a negation operation on $F\hat{}$ and fulfils to the properties:

$$c(f\hat{} \triangle g\hat{}) = c(f\hat{}) \bigtriangledown c(g\hat{}),$$

$$c(f\hat{} \bigtriangledown g\hat{}) \geq c(f\hat{}) \triangle c(g\hat{}).$$

If $'$ is a weak negation on $L$ then $c$ is a weak negation on $F\hat{}$, and if $'$ is an involution on $L$ then $c$ fulfils to:

$$c(c(c(f\hat{}))) = c(f\hat{}).$$

## 5 MODUS PONENS GENERATING FUNCTIONS ON $F\hat{}$

On $F\hat{}$ we can also generalize the concepts of $S$- and $R$-implication and for each of them the modus ponens generating functions mpgfS and mpgfR (Godo, Lopez de Mantaras et al. 1988; Hall 1990; Trillas, Valverde 1985; Valverde, Trillas 1985).

DEFINITION 1. A $S$-implication $->$: $F\hat{} \times F\hat{} \to F\hat{}$, a modus ponens generating function mpgfS: $F\hat{} \times F\hat{} \to F\hat{}$, a $R$-implication $=>$: $F\hat{} \times F\hat{} \to F\hat{}$ and a modus ponens generating function mpgfR: $F\hat{} \times F\hat{} \to F\hat{}$ are defined on $F\hat{}$ in the next way:

$$f\hat{} -> g\hat{} = c(f\hat{}) \bigtriangledown g\hat{},$$

$$\text{mpgfS}(f\hat{}, g\hat{}) = \inf\{h\hat{} \in F\hat{} \mid f\hat{} -> h\hat{} \geq g\hat{}\},$$

$$f\hat{} => g\hat{} = \sup\{h\hat{} \in F\hat{} \mid f\hat{} \triangle h\hat{} \leq g\hat{}\},$$

$$\text{mpgfR}(f\hat{}, g\hat{}) = f\hat{} \triangle g\hat{}.$$

THEOREM 3. For $m=$ mpgfR the next properties hold:

$$m(I\hat{}, I\hat{}) = I\hat{},\ m(0\hat{}, g\hat{}) = 0\hat{}, \tag{11}$$

$$\text{if } f\hat{} < h\hat{} \text{ and } 0\hat{} < g\hat{} \text{ then } m(f\hat{}, g\hat{}) < m(h\hat{}, g\hat{}), \tag{12}$$

$$\text{if } g\hat{} < h\hat{} \text{ and } 0\hat{} < f\hat{} \text{ then } m(f\hat{}, g\hat{}) < m(f\hat{}, h\hat{}), \tag{13}$$

$$m(f\hat{}, I\hat{}) < I\hat{} \text{ if } f\hat{} < I\hat{}, \tag{14}$$

$$m(f\hat{}, g\hat{}) < \min\{f\hat{}, g\hat{}\}, \text{ if } 0\hat{} < f\hat{}, g\hat{} < I\hat{}, \tag{15}$$

$$m(f\hat{}, g\hat{}) > 0\hat{} \text{ if } f\hat{}, g\hat{} > 0\hat{}. \tag{16}$$

PROOF. (11) follows from (7). (12)-(13) follow from (10). (14)-(15) follow from (9). (16) follows from (10), (7).

The properties (11)-(16) are based on properties of modus ponens generating functions discussed in (Hall 1990; Trillas, Valverde 1985; Valverde, Trillas 1985). We note here that (12) coincides with requirement of strict monotonicity of conclusions SMC and for $m=$ mpgfS only properties (11) and non-strict monotonicity in (12)-(13) are fulfiled.

## 6 EXAMPLE AND APPLICATIONS

Let expert system in medicine contains the rule:
If
HEMATURIA-INTENSITY=
MACROHEMATURIA
and
CLOT-OF-BLOOD= YES
and
CLOT-FORM= FORMLESS
then
HYPOTHESIS= UROLITHIASIS, $pv$= LARGE
and
HYPOTHESIS= TUMOR-OF-KIDNEY, $pv$= VERY-LARGE,
where plausibility values of rules, premises and facts are measured in linearly ordered scale with grades: {MINIMAL, VERY-SMALL, SMALL, AVERAGE, LARGE, VERY-LARGE, MAXIMAL }. And let the plausibility values of premises obtain the next values:
$pv$(HEMATURIA-INTENSITY=
MACROHEMATURIA) = MAXIMAL,
$pv$(CLOT-OF-BLOOD= YES) = VERY-LARGE,
$pv$(CLOT-FORM= FORMLESS) = LARGE.
Then the conclusions will obtain the next plausibility values:
$pv$(HYPOTHESIS= UROLITHIASIS)=
(LARGE, LARGE, VERY-LARGE),
$pv$(HYPOTHESIS= TUMOR-OF-KIDNEY)=
(LARGE, VERY-LARGE, VERY-LARGE).

Lexicographic comparison of plausibility values of hypotheses gives (HYPOTHESIS = TUMOR-OF-KIDNEY) as more plausible hypothesis while if we use only $\wedge$ (min) operation in conjunction and in modus ponens generating function then both hypotheses will obtain the same value LARGE.

An interpretation of $\wedge$-valuations of plausibility obtained on the output of expert system is very simple and gives an additional information about the number of plausible premises used in inference procedure and about plausibility values of these premises.

Some types of algebras of lexicographic valuations of plausibility are realized in expert systems shell LEXICO developed in expert systems laboratory of Kazan State Technological University. On the base of LEXICO different expert systems are constructed. For example hybrid expert system based on simulation model of chemical reactor was also created.

### Acknowledgements

The author is thankful to the referees for suggestions towards the improvement of the paper.

### References

A. V. Aho, J. E. Hopcroft, J. D. Ullman (1976). *The Design and Analysis of Computer Algorithms.* Massachusetts: Addison-Wesley.

I. Z. Batyrshin (1989). Lexicographical valuations in rule-based expert systems. In: *The Advent of AI in Higher Education.* Abstracts of Int. Symp. CEPES-UNESCO. Praga.

I. Z. Batyrshin (1990) Lexicographic valuations in expert systems. In: *Intellektualnye sistemy v zadachakh proektirovaniya, planirovaniya i upravleniya v usloviyakh nepolnoty informatsii.* Proceedings of All-Union symposium. Kazan, 133-136 (in Russian).

I.Z. Batyrshin (1993). Uncertainties with memory in decision-making and expert systems. In: *Proceedings of the Fifth IFSA World Congress'93.* Seoul, 737 - 740

I.Z. Batyrshin (1994) Lexicographic valuations of plausibility with universal bounds. *Izvestija RAN. Serija Tekhnicheskaja Kibernetika,* (in Russian) (submitted).

I. Z. Batyrshin, R. A. Zakuanov (1990a). On some generalization of Bellman-Zadeh approach to decision-making. In: *Towards a Unified Fuzzy Sets Theory. Abstracts of Third Joint IFSA-ES and EURO-WG Workshop on Fuzzy Sets.* Visegrad, 9-10.

I. Z. Batyrshin, R. Zakuanov (1993). Lexicographical valuations in decision-making and expert systems. In: *Proceedings of the First European Congress on Fuzzy and Intelligent Technologies EUFIT'93.* Aachen, 1599-1602.

S. Benferhat, C. Cayrol, D. Dubois, J. Lang, H. Prade (1993). Inconsistency management and prioritized syntax-based entailment. In: *13th Intern. Joint Conf. on Artificial Intelligence.* Chambery, 640-645.

D. Dubois, J. Lang, H. Prade (1990). A possibilistic assumption-based truth maintenance system with uncertain justifications, and its application to belief revision. In J.P. Martins, M. Reinfrank (ed.), *Truth Maintenagnce Systems.* Proceedings of ECAI-90 Workshop. Stockholm, 87-106.

D. Dubois, J. Lang, H. Prade (1992). Inconsistency in possibilistic knowledge bases. In L. A. Zadeh, J. Kacprzyk (eds.), *Fuzzy Logic for the Management of Uncertainty.* New York: John Wiley & Sons, 335-351.

H. Fargier, J. Lang, T. Schiex (1993). Selecting preferred solutions in fuzzy constraint satisfaction problems. In: *Proc.of First European Congress on Fuzzy and Intelligent Technologies.* Aachen, 3, 1128-1134.

P. C. Fishburn (1970). *Utility Theory for Decision Making.* New York: John Wiley & Sons.

R. Forsyth (1984) Fuzzy reasoning systems. In R. Forsyth (ed.), *Expert Systems. Principles and Case Studies,* London: Chapman and Hall, 51-62 (Russian translation).

L. L. Godo, R. Lopez de Mantaras, C. Sierra, A. Verdaguer (1988). Managing linguistically expressed uncertainty in MILORD application on medical diagnosis. *AICOM* 1, 14-31.

L. O. Hall (1990). The choice of ply operator in fuzzy intelligent systems. *Fuzzy Sets and Systems* 34, 135-144.

R. L. Keeney, H. Raiffa (1976). *Decisions with Multiple Objectives: Preference and Value Tradeoffs.* New York: John Wiley & Sons.

H. Moulin (1988). *Axioms of Cooperative Decision Making.* Cambridge: Cambridge University Press.

V. V. Podinovskii, V. M. Gavrilov (1975). *Optimization on sequentially used criteria.* Moscow: Sovetskoe Radio.

E. M. Reingold, J. Nievergelt, N. Deo (1977). *Combinatorial Algorithms. Theory and Practice.* New Jersey: Prentice-Hall.

E. Trillas, L. Valverde (1985). On mode and implication in approximate reasoning. In M.M. Gupta et al. (ed.), *Approximate Reasoning in Expert Systems.* Amsterdam: Elsevier Science.

L. Valverde, E. Trillas (1985). On modus ponens in fuzzy logic. In: *The International Symposium on Multiple-Valued Logic.* Kingston, Ontario, Canada.

E. Vilkas (1990). *Optimality in Games and Decisions.* Moscow: Nauka.

S. A. Weber (1983). A general concept of fuzzy connectives, negations and implications based on t-norms and t-conorms. *Fuzzy Sets and Systems* 11, 115-134.

L. A. Zadeh (1965). Fuzzy sets. *Information and Control* 8, 338-353.